\documentclass[11pt,letterpaper]{article}
\usepackage{setspace}
\usepackage[letterpaper,hmargin=1in,vmargin=1in]{geometry}

\usepackage{epsfig}
\usepackage{graphicx}
\usepackage{times}
\usepackage{amssymb}

\input{psfig.sty}

\usepackage{algorithmicx}
\usepackage[ruled]{algorithm}
\usepackage[noset]{algpseudocode}
\usepackage[noset]{algpascal}
\usepackage[noset]{algc}

\newenvironment{alginc}[1][pseudocode]{\medskip\algsetlanguage{#1}\begin{algorithmic}[1]}{\end{algorithmic}\medskip}

\newcommand\ASTART{\bigskip\noindent\begin{minipage}[c]{0.5\linewidth}}

\newcommand\AENDSKIP{\end{minipage}\bigskip}
\newcommand\AEND{\end{minipage}}

\def\final{0} 

\ifnum\final=1  

\newcommand{\vnote}[1]{[{\small vicky: \bf #1}]\marginpar{*}}
\newcommand{\nnote}[1]{[{\small navin: \bf #1}]\marginpar{*}}
\newcommand{\sidecomment}[1]{\marginpar{\tiny #1}}
\else 
\newcommand{\vnote}[1]{}
\newcommand{\nnote}[1]{}
\newcommand{\sidecomment}[1]{}

\newtheorem{lemma}{Lemma}[section]
\newtheorem{theorem}[lemma]{Theorem}
\newtheorem{definition}[lemma]{Definition}

\newtheorem{corollary}[lemma]{Corollary }

\newtheorem{claim}{Claim}

\newcommand{\ms}[1]{\ensuremath{\mathsf{#1}}}

\def\bull{\vrule height .9ex width .8ex depth -.1ex }
\newenvironment{proof}{\medbreak
\noindent
{\bf Proof.~}}{\unskip\nobreak\hfill\hskip 2em \bull \par\medbreak}

\newcommand{\PP}{P} 
\newcommand{\QQ}{Q} 
\newcommand{\DD}{\cal{D}} 
\newcommand{\DS}{{\cal{D}}_2} 
\newcommand{\DF}{{\cal{D}}_1} 
\newcommand{\DT}{{\cal{D}}_3} 
\newcommand{\diameter}[2]{\ms{diameter}(#1,#2)}
\newcommand{\BB}{B}
\newcommand{\CC}{C}
\newcommand{\TT}{T}
\newcommand{\EE}{E}
\newcommand{\II}{I}
\newcommand{\dist}[2]{||#1,#2||}
\newcommand{\RHT} {GHT}
\newcommand{\tolerant}{tolerant}

\newcommand{\HH}{H}
\newcommand{\distance}{\ms{dist}}
\newcommand{\LCP}{\ms{LCP}}

\newcommand{\DA} {{\sc Diheda} }

\newcommand{\eop}{\unskip\nobreak\hfill\hskip 2em \bull \par\medbreak}

\begin{document}
\sloppy

\title{An Efficient Approximation Algorithm for Point Pattern
Matching Under Noise\footnote{A preliminary version was presented at the 7th International Symposium,
  Latin American Theoretical Informatics (LATIN 2006)~\cite{ChGo2}.}}
\author{Vicky Choi \thanks{Corresponding author. Department of Computer Science, Virginia Tech,
    USA. vchoi@cs.vt.edu.} \and Navin Goyal \thanks{Department of Computer
    Science, McGill University, Canada. navin@cs.mcgill.ca.}
}

\maketitle

\begin{abstract}
Point pattern matching problems are of fundamental importance in various
areas including computer vision and structural bioinformatics. In this
paper, we study one of the more general problems,
known as LCP (largest common point set problem):
Let $\PP$ and $\QQ$ be two point sets in $\mathbb{R}^3$, and
let $\epsilon \geq 0$ be a tolerance parameter, the problem is to find a
rigid motion $\mu$ that maximizes the cardinality of
subset $\II$ of $Q$, such that
the Hausdorff distance $\distance(\PP,\mu(\II)) \leq
\epsilon$.  We denote the size of the optimal solution to the above problem
by $\LCP(P,Q)$. 
The problem is called exact-LCP for $\epsilon=0$,  and 
\tolerant-LCP when $\epsilon>0$ and the minimum interpoint distance 
is greater than $2\epsilon$.
A $\beta$-distance-approximation algorithm for tolerant-LCP finds a subset $I
\subseteq \QQ$
such that $|I|\geq \LCP(P,Q)$ and $\distance(\PP,\mu(\II)) \leq
\beta \epsilon$ for some $\beta \ge 1$.

This paper has three main contributions.
(1) We introduce a new algorithm, called {\DA}, which
gives the fastest known deterministic $4$-distance-approximation algorithm for
\tolerant-LCP.  
(2) For the exact-LCP, when the matched set is required to be large, 
we give a simple sampling strategy that 
improves the running times of all known deterministic algorithms, 
yielding the fastest known deterministic
algorithm for this problem. 
(3) We use expander graphs to speed-up the
  \DA algorithm for \tolerant-LCP when the size of the matched 
set is required to be large, at the expense of 
   approximation in the matched set size.
Our algorithms also work when the transformation 
$\mu$ is allowed to be scaling transformation.

{\bf Keywords.} Point Pattern Matching, Largest Common Point Set
\end{abstract}

\section{Introduction}

The general problem of finding large similar common substructures in
two point sets arises in many areas ranging from computer
vision to structural bioinformatics. 
In this paper, we study one of the more general problems, 
known as
the \emph{largest common point set problem} (LCP), which has several variants
to be discussed below.                    

\paragraph{Problem Statement.} 
Given two point sets in $\mathbb{R}^3$, $\PP=\{p_1, \ldots, p_m\}$  and 
$\QQ=\{q_1, \ldots, q_n\}$, and an error parameter
$\epsilon \geq 0$, we want to find a
rigid motion $\mu$ that maximizes the cardinality of subset $\II
\subseteq \QQ$, such that  
$\distance(\PP, \mu(\II)) \leq \epsilon$. For an optimal set $\II$, denote $|\II|$ by $\LCP(P,Q)$. 
There are two commonly used distance measures between point sets:
\emph{Hausdorff distance} 
and \emph{bottleneck distance}.
The Hausdorff distance $\distance(P,Q)$ between two point sets $P$ and
$Q$ is given by $\max_{q\in Q}\min_{p \in P}||pq||$.
The bottleneck distance $\distance(P,Q)$ between two point sets $P$
and $Q$ is given by $\min_{f}\max_{q \in
Q}||f(q) - q ||$, where $f:Q\rightarrow P$ is an injection.
Thus we get two versions of the LCP depending on which
distance is used. 

Another distinction that is made is between the \emph{exact}-LCP and
the \emph{threshold}-LCP. In the former we have $\epsilon = 0$ and in
the latter we have $\epsilon > 0$. The exact-LCP is computationally
easier than the 
threshold-LCP; however, it is not useful when the data suffers from
round-off and sampling errors, and when we wish to measure the
resemblance between two point sets and do not expect exact
matches. These problems are better modeled by the threshold-LCP,
which turns out to be harder, and various kinds of approximation
algorithms have been considered for it in the literature (see below).
A special kind of threshold-LCP in which one assumes that 
the minimum interpoint distance 
is greater than the error parameter $2\epsilon$ is called
\emph{\tolerant}-LCP. 
\tolerant-LCP  more accurately captures many problems arising in practice,
and  it appears that it is algorithmically easier than threshold-LCP.
Notice that for the \tolerant-LCP, 
the Hausdorff and bottleneck distances are essentially the same in the
sense that the
problem has a solution of Hausdorff distance $\le \epsilon$ if and only if
the solution is of bottleneck distance $\le \epsilon$.
Thus, for the \tolerant-LCP, there is no need to specify
which distance is in use.

In practice, it is often the case that the size of the solution set $\II$ to the
LCP is required to be at least a certain fraction of the minimum of
the sizes of the two point sets: 
$|\II| \geq \frac{1}{\alpha} \min(|\PP|, |\QQ|),$
where $\alpha$ is a positive constant. This version of the LCP is
known as the $\alpha$-LCP. 
A special case of the LCP which requires matching the entire set $Q$ is
called {\em Pattern Matching} (PM) problem. Again, we have exact-PM,
threshold-PM, and tolerant-PM versions.

In this paper, we focus on approximation algorithms for \tolerant-LCP
and tolerant-$\alpha$-LCP.
There are two natural notions of approximation. (1) \emph{Distance approximation:} The 
algorithm finds a transformation that
brings a set $\II \subseteq \QQ$ of size at least $\LCP(\PP, \QQ)$
within distance $\epsilon'$ for some constant $\epsilon' >
\epsilon$. (2) \emph{Size-approximation:} The algorithm guarantees that
$|\II| \geq (1-\delta) \LCP(\PP,\QQ)$, for constant $\delta \in [0,1)$. 

\paragraph{Previous work.} The LCP has been extensively investigated
in  computer vision (e.g. \cite{Olson97}), 
computational geometry
(e.g. \cite{Alt88}), and also
finds applications in computational structural biology (e.g. \cite{Wolfson2}).
For the exact-LCP problem, there are four simple and popular
algorithms: {\em alignment} (e.g. \cite{IrRa96,Akutsu2}), 
{\em pose clustering} (e.g. \cite{Olson97}), {\em geometric
  hashing} (e.g. \cite{LaWo})  and  
{\em generalized Hough transform} (GHT) (e.g. \cite{HeBo}).
These algorithms are often confused with one another in the
literature.  
For convenience of the reader, we include brief
descriptions of these algorithms in the appendix.
Among these
four algorithms, the most efficient algorithm is GHT.

\paragraph{Exact algorithms for \tolerant-LCP.} 
As we mentioned above, the \tolerant-LCP (or more generally,
 threshold-LCP) is a better model of many situations that arise in
 practice. 
However, it turns out that it is considerably more difficult to solve 
the tolerant-LCP than the exact-LCP.  
Intuitively, a fundamental difference between
the two problems lies in the fact that for the exact-LCP the set of rigid motions,
that may potentially correspond to the solution, is {\em discrete} and
can be easily enumerated.  Indeed, the
algorithms for the exact-LCP are all based on the (explicit or
implicit) enumeration of 
rigid motions that can be obtained by matching triplets to triplets.
On the other hand, for the tolerant-LCP this set is {\em continuous},
and hence the direct enumeration strategies do not work.
Nevertheless, the optimal rigid motions can be 
characterized by a set of high degree polynomial equations as in
\cite{AmChGa}.
A similar characterization was
made by Alt and Guibas in \cite{AltGuibas} for the 2D tolerant-PM
problem and by the authors in \cite{ChGo} for the 3D tolerant-PM. 
All known algorithms for the threshold-LCP use these characterizations
and involve solving systems of high degree equations which leads 
to  ``numerical instability problem''~\cite{AltGuibas}.
Note that exact-LCP and the exact solution for tolerant-LCP are two
distinct problems. (Readers are cautioned not to confuse these two
problems as in Gavrilov et al.~\cite{Gavrilov}.) 
Amb{\"u}hl et al.~\cite{AmChGa} gave an algorithm for tolerant-LCP
with running time $O(m^{16}n^{16}\sqrt{m+n})$. The algorithm in
\cite{ChGo} for threshold-PM can be
adapted to solve the tolerant-LCP in $O(m^6n^6 (m+n)^{2.5})$
time. Both algorithms are for bottleneck distances. These algorithms
can be modified to solve threshold-LCP under Hausdorff distance  with a better running
time by replacing the maximum bipartite graph matching algorithm which runs in
$O(n^{2.5})$ with the $O(n\log n)$ time algorithm for nearest neighbor search.
Both of these algorithms are for the general threshold-LCP, but
to the best of
our knowledge, these algorithms are the only known exact algorithms
for the \tolerant-LCP also.

\paragraph{Approximation algorithms for \tolerant-LCP.} 
Like threshold-LCP, the exact algorithm for
threshold-PM is difficult, even in 2D (see \cite{AltGuibas}). 
Two types of approximation algorithms were studied.
First,
Goodrich~et~al~\cite{GoMiOr} showed that
there is a small discrete set of rigid motions which contains a rigid motion
approximating (in distance) the optimal rigid motion for the
threshold-PM problem, and thus the threshold-PM problem
can be solved approximately by an enumeration strategy.
Based on this idea and the alignment approach of enumerating all
possible such discrete rigid motions, 
Akutsu~\cite{Akutsu1}, and 
Biswas and Chakraborty~\cite{ChBi,BiCh} gave
distance-approximation algorithms with
running time $O(m^4n^4 \sqrt{m+n})$ for the threshold-LCP under bottleneck
distance, which can be modified to give $O(m^3n^4 \log{m})$ time
algorithm for the tolerant-LCP.
Second,
Heffernan and Schirra~\cite{HeSc} introduced approximate {\em decision}
algorithms to approximate the minimum Hausdorff distance between two point sets.
Given $\epsilon >0$, their algorithm answers correctly (YES/NO) if $\epsilon$
is not too close to the optimal value $\epsilon^*$ (which is the minimum
Hausdorff distance between the two point sets) and DON'T KNOW if the
answer is too close 
to the optimal value. 
Notice that this approximation framework can not be ``similarly'' adopted 
 to the LCP problem because in the LCP case there are two parameters
 -- size and distance -- to be optimized.
 This appears to be mistaken by Indyk et al. in \cite{Indyk,Gavrilov}
 where their approximation algorithm for tolerant-LCP is not well defined.
Cardoze and Schulman~\cite{CaSc} gave an approximation algorithm (with possible false
positives) but the transformations are restricted to 
translations for the LCP problem.
Given $\alpha$, let $\epsilon_{min}(\alpha)$ denote the smallest 
$\epsilon$ for which
$\alpha$-LCP exists; given $\epsilon$, let 
$\alpha_{min}(\epsilon)$ 
denote the smallest $\alpha$ for which $\alpha$-LCP exists.
Biswas and Chakraborty~\cite{ChBi,BiCh}
combined the idea from Heffernan and Schirra and
the algorithm of Akutsu~\cite{Akutsu1} to give a size-approximation algorithm
which returns  $\alpha_u>\alpha_l$ such that $\min\{\alpha: \epsilon > 8
\epsilon(\alpha)\} \ge \alpha_u \ge \alpha_{min}(\epsilon)$ and
$\alpha_{min}(\epsilon) > \alpha_l \ge \max\{\alpha: \epsilon <
\frac{1}{8}\epsilon_{min}(\alpha)\}$.
However, all these approximation algorithms still take high running
time of $\tilde{O}(m^3 n^4)$ (the notation $\tilde{O}$ hides poly log
factors in $m$ and $n$).

\paragraph{Heuristics for \tolerant-LCP.} 
In practice, the tolerant-LCP is solved heuristically by using the  
geometric hashing and GHT algorithms for which
rigorous analyses are only known for the exact-LCP. 
For example, the algorithms in~\cite{Rapid97,Olson97} are for tolerant-LCP but 
the analyses are for exact-LCP only.
Because of its practical
performance, the exact version of GHT was
carefully analyzed by Akutsu~et~al.~\cite{Akutsu2}, 
and a randomized version of the exact version of 
geometric hashing in 2D was given by Irani and Raghavan~\cite{IrRa96}.
The tolerant version of GHT (and geometric hashing) is based on the
corresponding exact version by replacing the exact matching with the
approximate matching which requires a distance measure to compare the
keys.  We can no longer identify the optimal rigid motion by the 
maximum votes as in the exact case. Instead, the tolerant
version of GHT \emph{clusters} the rigid motions (which are points in
a six-dimensional space) and
heuristically approximates the optimal rigid motion by a
rigid motion in the largest cluster. Thus besides not giving any
guarantees about the solution, this heuristic requires clustering in
six dimensions, which is computationally expensive.
  

\paragraph{Other Related Work.} There is some closely related
work that aims at computing the {\em minimum} Hausdorff distance for
PM (see, e.g., \cite{Kedem} and references therein). Also, the problems we are considering
can be thought of as the point pattern matching problem under {\em uniform
distortion}. Recently, there has been some work on point pattern
matching under {\em non-uniform distortion}~\cite{Kenyon,Akutsu3}.

\paragraph{Our results.} 
There are three results in this paper.
First, we introduce  a new  distance-approximation algorithm for
\tolerant-LCP algorithm, called {\DA} (because our algorithm is based on
	      {\sc Dihedral Angle} comparisons).

\begin{theorem}\label{thm:main}
Let $\PP, \QQ \in \mathbb{R}^3$ of size $m$ and $n$, with $m \ge n$, and
$\epsilon>0$. 
Suppose that 
interpoint distances in $\PP$ and in $\QQ$ be $>2 \epsilon$ (this is
the condition for \tolerant-LCP).
\DA (see Algorithm ~\ref{TH-approximate}) 
 finds a rigid motion $\mu$ and a subset
 $\II$ of $\QQ$ such that
 \begin{itemize}
   \item $|I| \ge \LCP(\PP,\QQ)$ and
     \item  $\distance(\PP,\mu(\II)) \leq
4\epsilon$
 \end{itemize}
in $O(m^3n^3 \log{m})$ time. 
\end{theorem}

\DA is simple and more efficient than the known
 distance-approximation algorithms (which are alignment-based) for
\tolerant-LCP. 
The running time of \DA is $O(m^3n^3 \log{m})$ in the worst case.
For general input, we expect the algorithm to be much faster
because it is simpler and more efficient than the
previous heuristics that are known to be fast in practice. 
This is because our clustering step is simple (sorting 
linearly ordered data) 
while the clustering step in those
heuristics requires clustering high-dimensional data.

Second, based on a combinatorial observation, we improve the 
algorithms for exact-$\alpha$-LCP  by a linear factor for pose
clustering or GHT
and a quadratic factor for alignment or geometric hashing.
This also corrects a mistake by Irani and Raghavan~\cite{IrRa96}. 

Finally, we achieve a similar speed-up for \DA using a sampling approach
  based on expander graphs
  at the expense of 
   approximation in the matched set size.
We remark that this result
is mainly of theoretical interest because of the large constant factor involved.
Expander graphs have been used before in geometric optimization for
   fast deterministic algorithms \cite{AjMe,KaSh97}; however, the way
   we use these graphs appears to be new.
Our results also hold when we extend the set of transformations to
scaling; for simplicity we restrict ourselves to rigid
motions in this paper. 

\paragraph{Outline.} 
The paper is organized as follows.  The rest of this section contains
some preliminaries.  
In Section 2 we introduce our new distance-approximation algorithm
for tolerant-LCP. 
In Section 3 we show how a simple deterministic sampling
strategy based on the pigeonhole principle yields speed-ups 
for the exact-$\alpha$-LCP algorithms.
In Section 4 we show how to use expander graphs to further speed up the
  \DA algorithm for \tolerant-$\alpha$-LCP  at the expense of 
   approximation in the matched set size.
Section 5 is the conclusion.
In the appendix, we recall and compare the
   existing four basic algorithms for exact-LCP: {\em
   pose clustering}, {\em alignment}, {\em GHT} 
and {\em geometric hashing}.

\begin{figure}[h]
    \begin{center}
      \centerline{\psfig{figure=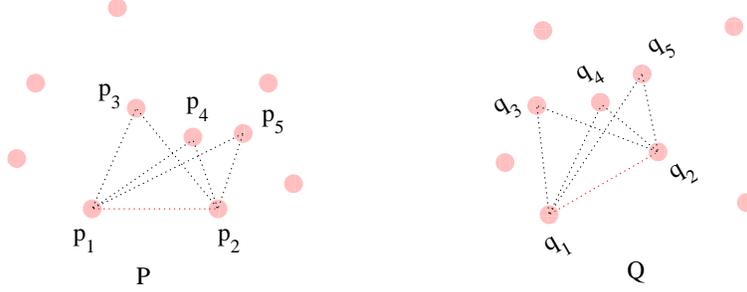,height=1.5in}}
      \caption{
      In this example, the rigid motion is obtained by matching $C_Q=\{q_1, q_2, \ldots, q_5\}$ in $\QQ$ to 
 $C_P=\{p_1, p_2, \ldots, p_5\}$ in $\PP$.
We have $\LCP(\PP,\QQ)=|C_P|=|C_Q|=5$.
The corresponding 5-matching transformation $\mu$ can be discovered by 
matching $(q_1,q_2)$ to $(p_1,p_2)$, the rigid motions $\mu_i$ that
transform $(q_1,q_2, q_i)$ to $(p_1,p_2, p_i)$ for $i=3,4,5$
are all the same and thus $\mu=\mu_3=\mu_4=\mu_5$ will get 3 votes, 
which is the maximum.
}
      \label{Fig:basis-example}
    \end{center}
  \end{figure}

\paragraph{Terminology and Notation.} 
For a transformation $\mu$, denote by $I_{\mu}$ the set of points in
$\mu(Q)$ that are within distance $\epsilon$ of some point in $\PP$. 
We call $I_{\mu}$ the matched
set of $\mu$ and say that 
$\mu$ is an $|I_{\mu}|$-matching. 
We call the 
transformation 
$\mu$ that maximizes $|I_{\mu}|$ the {\em maximum matching} transformation.
A {\em basis} is a minimal (for containment relation) ordered tuple of points which is required to
uniquely define a rigid motion.
For example, in 2D every ordered pair is a basis; while in 3D, every
non-collinear triplet is a basis. 
In Figure~\ref{Fig:basis-example}, a rigid motion in 3D is specified by mapping a
basis $(q_1,q_2,q_3)$ to another basis $(p_1,p_2,p_3)$. 
We call a key used to represent an ordered tuple $S$ a {\em rigid motion
  invariant} key if it satisfies the following:
(1) the key remains the same for all $\mu(S)$ where $\mu$ is any
rigid motion, and (2) for any two ordered tuples $S$ and $S'$ with
the same rigid motion invariant key there is a unique rigid motion
$\mu$ such that $\mu(S)=S'$.
For example, as rigid motion preserves orientation and
 distances among points, given a non-degenerate triangle
$\Delta$, the 3 side lengths of $\Delta$ together with the
orientation (the sign of the determinant of the ordered triplet) form a rigid
motion invariant key for $\Delta$ in $\mathbb{R}^3$.  Henceforth, for
  simplicity of exposition, in
  the description of our algorithms we will omit the orientation
  part of the key.

\section{\DA}
In this section, we introduce a new distance-approximation algorithm, 
called {\em \DA}, for \tolerant-LCP.
The algorithm
is based on a simple geometric observation. 
It can be seen as an improvement of a
known GHT-based heuristic such that the output has theoretical
guarantees.

\subsection{Review of GHT}
First, we review the idea of the pair-based version of GHT for
exact-LCP. See the
appendix or ~\cite{Akutsu2,Olson97} for more details.
For each congruent pair, say $(p_1,p_2)$ in $\PP$ and $(q_1,q_2)$ in 
$\QQ$,  and for each of the remaining points $p \in \PP$ and $q \in
\QQ$, if $(q_1,q_2, q)$ is congruent to  
$(p_1,p_2, p)$,  compute the rigid motion $\mu$ 
 that matches $(q_1,q_2, q)$ to 
$(p_1,p_2, p)$. 
We then cast one vote for $\mu$. The rigid motion that receives the
maximum number of votes corresponds to the maximum matching transformation
sought. 
See Figure~\ref{Fig:basis-example} for an example.

\subsection{Comparable rigid motions by dihedral angles}
For the exact-LCP, one only needs to compare rigid motions by equality
(for voting). For the \tolerant-LCP, one needs to measure 
how close two rigid motions are. In $\mathbb{R}^3$, each rigid motion can be described by 6 parameters 
(3 for translations and 3 for rotations).  
How to define a distance measure between rigid motions?
We will show below that
the rigid motions considered in our algorithm are
related to each other in a simple way that enables a natural notion
of distance between the rigid motions. 

\begin{figure}[h]
    \begin{center}
	\centerline{\psfig{figure=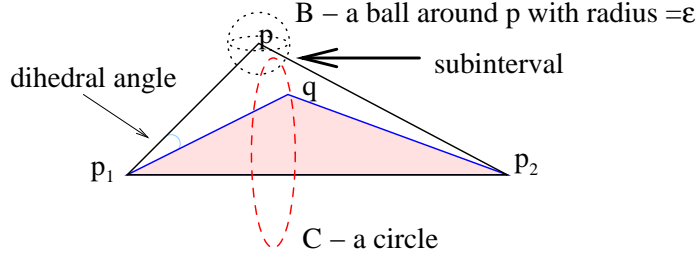,height=1.3in}}
      \caption{The dihedral angle is the angle between planes formed by
	$(p_1,p_2,q)$ and $(p_1,p_2,p)$.
	The rotation angles of transformations that rotate $q$ about
	$\stackrel{\longrightarrow}{p_1p_2}$
 to within
$\epsilon$ of $p$ form a subinterval of $[0,2\pi)$.}
      \label{Fig:dihedral-angle}
    \end{center}

  \end{figure}

\paragraph{Observation.} In the pair-based version of GHT as described
above, 
the rigid motions to be compared have a special property: 
the rigid motions transform a common pair --- 
they all match $(q_1, q_2)$ to $(p_1,p_2)$ in Figure~\ref{Fig:basis-example}. 
Two such transformations no longer differ in all 6 parameters but differ
in only one parameter. To see this, we first recall that 
a dihedral angle is the angle between
two intersecting planes; see Figure~\ref{Fig:dihedral-angle} for an example.
In general, we can decompose the rigid motion for matching $(q_1, q_2,
q_3)$ to $(p_1, p_2, p_3)$ into two parts: first, we transform $(q_1,
q_2)$ to $(p_1,p_2)$ by a transformation $\phi_1$; then we rotate the
point $\phi_1(q_3)$ about 
	$\stackrel{\longrightarrow}{p_1p_2}$
by an angle
$\theta$,  where $\theta$ is the dihedral angle between the planes
$(p_1,p_2,p_3)$ and $(\phi_1(q_1),\phi_1(q_2),\phi_1(q_3))$. 
This will  bring $q_3$  to coincide with $p_3$. Thus, we have the
following lemma: 
\begin{lemma}
Let $(p_1, p_2, p_3)$ and $(q_1, q_2, q_3)$ be two congruent
non-collinear triplets,
and let  $\phi_1$ be a rigid motion that takes $q_i$ to $p_i$ for
$i=1,2$. Let $\phi_2$ be the rotation about 
	$\stackrel{\longrightarrow}{p_1p_2}$
by an angle $\theta$,
 where $\theta$ is the dihedral angle between the planes
$(p_1,p_2,p_3)$ and $(\phi_1(q_1),\phi_1(q_2),\phi_1(q_3))$. 
Then the unique rigid motion that takes   
$(p_1, p_2, p_3)$ to $(q_1, q_2, q_3)$ is equal to $\phi_2 \circ \phi_1$.
\end{lemma}
We now state another lemma that will be useful in the description and
proof of correctness of \DA.
Let $(p_1,p_2,p)$ and $q$ be four points as shown in
Figure~\ref{Fig:dihedral-angle}. 
Consider the rotations about 
	$\stackrel{\longrightarrow}{p_1p_2}$
that take $q$ to within
$\epsilon$ of $p$. The rotation angles of these
transformations form a
subinterval 
of $[0,2\pi)$. 
This is because a circle $C$ 
(corresponding to the trajectory of $p$) 
intersects with the sphere $B$ (around $p$ with 
radius $\epsilon$) at at most two points (corresponding to a subinterval of 
$[0,2\pi)$), as shown in Figure \ref{Fig:dihedral-angle}.
That is, we have the following lemma:
\begin{lemma}\label{lem:subinterval}
Let $p_1, p_2, p, q \in \mathbb{R}^3$ be four points (not necessarily non-collinear), 
then the rotation angles of transformations 
that rotate $q$ 
about $\stackrel{\longrightarrow}{p_1p_2}$
to within $\epsilon$ of $p$ form a subinterval of $[0, 2\pi)$.
\end{lemma}

\subsection{Approximating the optimal rigid motion by the ``diametric''
      rigid motion}

For a point set  $S \subset \mathbb{R}^3$,  we call a pair of points
$\{p,q\} \in S^2$ \emph{diameter-pair} if $||p-q|| =
\ms{diameter}{(S)}$.
A rigid motion of $\QQ$ that takes $q_1$ to $p_1$ and $q_2$
 on the
line $p_1p_2$ and closest possible to $p_2$ is called a
$(p_1,p_2,q_1,q_2)$-rigid motion.
Based on an idea similar to the one behind Lemma~2.4 in 
Goodrich et al.~\cite{GoMiOr}, we have the following lemma:
\begin{lemma}\label{lem:diametric-lemma}
Let $\mu$ be a rigid motion such that each point of $\mu(S)$,
where $S \subseteq \QQ$, is within
distance $\epsilon$ of a point in $\PP$. 
Let $\{q_1,q_2\}$ be a diameter-pair of $S$. 
Let $p_i \in \PP$ 
be the closest point to $\mu(q_i)$ for $i=1,2.$ 
Then we have a $(p_1,p_2,q_1,q_2)$-rigid motion $\mu'$ of $\QQ$ such
that each point of $\mu'(S)$ is within $4\epsilon$ of a point in
$\PP$. 
\end{lemma}

\noindent{\em Proof Sketch.} 
Translate $\mu(q_1)$ to $p_1$; this translation shifts  each point by at most
$\epsilon$. Next, rotate about $p_1$ 
such that  $\mu(q_2)$ is closest to $p_2$ (which implies
$\mu'(q_1),\mu'(q_2)$ and $p_2$ are collinear). 
Since $\{q_1, q_2\}$ is a diameter-pair, this rotation moves each point
by at most $2\epsilon$. Thus, each point is at most $\epsilon + \epsilon
+ 2\epsilon = 4\epsilon$ from its matched point. 
\eop

\subsection{Approximation algorithm for \tolerant-LCP}
We first describe the idea of our algorithm \DA.  
Input is two point sets in $\mathbb{R}^3$, $\PP=\{p_1, \ldots, p_m\}$  and 
$\QQ=\{q_1, \ldots, q_n\}$ with $m \ge n$,  and $\epsilon \geq 0$.  
Suppose that the optimal rigid motion $\mu_0$ was
achieved by matching a set $I_{\mu_0} = \{q_1, q_2,
\ldots, q_k\} \subseteq \QQ$ to $J_{\mu_0} = \{p_1, p_2, \ldots, p_k\}
\subseteq \PP$. WLOG, assume that $\{q_1,q_2\}$  is the diameter pair
of $I_{\mu_0}$.
Then by Lemma~\ref{lem:diametric-lemma}, there exists a 
$(p_1,p_2,q_1,q_2)$-rigid motion $\mu$ of $\QQ$ such that 
$\mu(I_{\mu_0})$ is within $4\epsilon$ of a point in
$\PP$. 
Since we do not know the matched set, we do not know a diameter-pair
for the matched set either. 
Therefore, we exhaustively go through each possible
pair.
Namely, for each pair $(q_1,q_2) \in \QQ$ and each pair $(p_1,p_2) \in \PP$, if they
are approximately congruent then we find a $(p_1,p_2,q_1,q_2)$-rigid
motion $\mu$ of $\QQ$ that matches as many remaining points as
possible.  
Note that $(p_1,p_2,q_1,q_2)$-rigid motions are 
determined up to a rotation about the line $p_1p_2$.  
By Lemma~\ref{lem:subinterval}, the rotation angles that bring
$\mu(q_i)$ to within $4\epsilon$ of $p_i$  form a subinterval of
$[0,2\pi)$. 
And the number of non-empty intersection subintervals
corresponds to the size of the matched set. 
Thus, to find $\mu$, for each pair $(p,q) \in \PP\setminus\{p_1,p_2\} \times
\QQ\setminus\{q_1,q_2\}$, we compute the dihedral angle interval
according to Lemma~\ref{lem:subinterval}. 
The rigid motion $\mu$ sought corresponds to an angle $\phi$ that lies
in the maximum number of dihedral intervals.
The details of the algorithm are described in Algorithm \ref{TH-approximate}.

\begin{small}
\begin{algorithm}
\caption{\DA}\label{TH-approximate}
\begin{alginc}

\Procedure{Preprocessing}{}
\For {each pair $(p_1,p_2)$ of $\PP$}
\State Compute and insert the key of $||p_1p_2||$ into a dictionary $\DF$;
\EndFor

\For {each triplet $(p_1,p_2,p_3)$ of $\PP$}
\State Compute and insert the {\em rigid motion invariant} key for 
$(p_1,p_2,p_3)$ into a dictionary $\DS$;
\EndFor 

\EndProcedure

\Procedure{Recognition}{}
\For {each pair $(q_1,q_2)$ of ${Q \choose
    2}$ }
\Comment{This can be reduced by the edge set of an expander of $Q$.}
\If {$[||q_1q_2||-2\epsilon, ||q_1q_2||+2\epsilon]$ exists in $\DF$}
\State Initialize an empty dictionary $\DT$ of pairs; 
\For {each remaining point $q \in \QQ$}
\State Compute and search the range $[||q_1q_2||-2\epsilon,
  ||q_1q_2||+2\epsilon] \times [||qq_1||-2\epsilon, ||qq_1||+2\epsilon]
\times [||qq_2||-2\epsilon, ||qq_2||+2\epsilon]$  of
$(q_1,q_2,q)$ in $\DS$;
\Comment{e.g. using a kd-tree.}
\For {each entry $(p_1,p_2,p)$ found} 
\State If {$(p_1,p_2)$ exists in $\DT$,} increase its vote; otherwise insert  
$(p_1,p_2)$ into $\DT$ with one vote;
\State Append the matched pair $(q,p)$ to the list associated with $(p_1,p_2)$;
\EndFor
\EndFor
\EndIf
\Comment Compute the  maximum transformation that matches $(q_1,q_2)$
to $(p_1,p_2)$.
\For {each pair $(p_1,p_2)$ in the dictionary $\DT$}
\State Compute a transformation $\phi$ that brings $q_1$ to $p_1$ and
$q_2$ closest to $p_2$;
\State For each matched pair $(q,p)$ of the associated list of $(p_1,p_2)$,
compute an interval of dihedral angles such that
 $\phi(q)$ is within $4 \epsilon$ of $p$;
\State Sort all the intervals of dihedral angles; and find a dihedral angle $\psi$ that
occurs in the largest number $V$ of intervals; 

 \State Compute the transformation $\mu$ by the composition of $\phi$ and the
 rotation about $p_1p_2$ by angle $\psi$;
\Comment $\mu$ brings $V+2$ points of $Q$ to within $4 \epsilon$ of
 some matched points in $P$.
\State Keep the maximum matched set size and the corresponding transformation;
\EndFor
\EndFor
\EndProcedure
\end{alginc}
\end{algorithm}
  
\end{small}

\noindent{\bf Time Complexity.} 
For each triplet in $Q$, using kd-tree for range query,
it takes $O(m^{3\cdot(1 - \frac{1}{3})} + m^3
+ m^3 \log{m^2}) = O(m^3\log{m})$ for lines 11--20.
For each pair $(q_1,q_2)$ and $(p_1,p_2)$, we spend time $O(mn)$ to
find the subintervals for the dihedral angles, and time
$O(mn \log{m})$ to sort these subintervals and do the scan to find an
angle that lies in the maximum number of subintervals. Thus the total
time is $O(m^3n^3 \log{m})$.

\section{Improvement by pigeonhole principle}
In this section we show how a simple deterministic sampling
strategy based on the pigeonhole principle yields speed-ups 
for the four basic algorithms for exact-$\alpha$-LCP.  
Specifically, we get a linear speed-up for pose clustering and \RHT, 
and quadratic speed-up for alignment and geometric hashing.
It appears to have been erroneously concluded previously that no such
improvements were possible deterministically 
\cite{IrRa96}.

In pose clustering or \RHT,  suppose we know a pair $(q_1, q_2)$ in $\QQ$
that is in the sought matched set, then 
the transformation sought will be the one receiving the maximum number
of votes among the transformations
computed for
$(q_1,q_2)$.  
Thus if we have chosen a pair
$(q_1,q_2)$ that lies in the matched set, then the maximum matching
transformation will be found.  
We are interested in the question  ``can we 
find a pair in the matched set  without exhaustive enumeration''? The answer is
yes: 
we only need to try a linear number of pairs
$(q_1,q_2)$ to find the maximum matching transformation or conclude that
there is none that matches at least $\frac{n}{\alpha}$ points. 


We are given a set $\QQ = \{q_1, \ldots, q_n\}$, and let $\II
 \subseteq \QQ$ be an unknown set of size $ \geq \frac{n}{\alpha}$ for
 some constant $\alpha>1$.  We need to discover a pair $(p,q)$ with
  $p,q \in \II$ by using queries of the following type.  A query
 consist of a pair $(a,b)$ with $a,b \in \QQ$.  If we have $a,b \in
 \II$, the answer to the query is YES, otherwise the answer is NO.  Thus our goal
 is to devise a deterministic query scheme such that as few queries
 are needed as possible in the worst case (over the choice of $\II$)
 before a query is answered YES.  
Similarly, one can ask the question about querying triplets to discover a triplet
 entirely in $\II$.
\begin{theorem}\label{thm:pigeonhole}
For an unknown set $\II \subseteq \QQ$ with $|\II| \geq \frac{n}{\alpha}$ and $|\QQ|=n$ 
using queries as described above, 
\begin{description}
\item(1) it suffices to query  $O(\alpha n)$ pairs to discover a pair  in $\II$;
\item(2) it suffices to query  $O(\alpha^{2} n)$ triplets to discover a
  triplet in $\II$.
\end{description}
\end{theorem}

\begin{proof} The proof is based on the pigeonhole
  principle. To prove (1),
we assume for simplicity that $\alpha$ and $n\over\alpha$
are both integers.  Partition the set $\QQ$ into $\frac{n}{\alpha}$ subsets of
size $\alpha$ each. Since the size of $\II$ is more than 
$\frac{n}{\alpha}$,
by the pigeonhole principle, there is a pair of points
in $\II$ that lies in one of the above chosen subsets. Thus
querying all pairs in these subsets will discover $\II$. This gives
that $\frac{n}{\alpha}{\alpha \choose 2} \sim \alpha n$ queries are 
sufficient to discover $\II$.

Similarly, to prove (2), partition $\QQ$ into
$n\over {2\alpha}$ subsets $P_1, \ldots, P_{n\over {2\alpha}}$ of size $2\alpha$ each (we
assume, as before, that $2\alpha$ and $n\over {2\alpha}$ are both integers).
Now we test all triplets that lie in the $P_i$'s. Any set 
$I \subseteq \QQ$ that intersects with each of the $P_i$'s in at
most $2$ points has size $\leq {n\over \alpha}$. Hence if $|I|> {n \over \alpha}$ then it must intersect
with one of the sets above in at least $3$ points. Thus testing the
triplets from the $P_i$'s is sufficient to discover $\II$.
The number of triplets tested is $\frac{n}{2\alpha}{2\alpha \choose 3} \sim \alpha^2n$. 
\end{proof}

Remark: It can be shown that the schemes in the proof above are the best possible in requiring
the smallest number of queries (up to constant factors).

In alignment and geometric hashing algorithms if we have chosen a triplet
 $(q_1,q_2,q_3)$ from the maximum matching set $\II \subseteq
\QQ$ then we will discover $\II$.
The question, as before, is how many triplets in
$\QQ$ need to be queried to discover a set $I$ of size $>{n\over \alpha}$.
By Theorem~\ref{thm:pigeonhole} (2), we only need to query $O(\alpha^{2} n)$
triplets.
Thus the running times of both alignment and geometric hashing are
improved by a factor of $\Theta(n^2)$. 

See Table \ref{voting-time-comp} for the time complexity comparison of deterministic
algorithms for exact-$\alpha$-LCP in $\mathbb{R}^3$.

Finally, our approximation algorithm for tolerant-LCP adapts naturally
for exact-$\alpha$-LCP with pigeonhole sampling.  \nnote{We first need
  to explain the algorithm for this case and then analyze it.}
We analyze the running time of our algorithm for exact-$\alpha$-LCP
 with the pigeonhole sampling
of pairs.  In the
exact case, each exact matched pair of points $(q,p)$ 
corresponds to a single dihedral angle. We thus find the dihedral
angle that occurs the maximum number of times by sorting all the
dihedral angles. For a fixed pair $(q_1,q_2)$ and a point $q$ in $\QQ$ the
number of triplets $((p_1,p_2),p_3)$ in $\PP$ that match $((q_1,q_2),q_3)$ is
bounded above by $3\HH_2(m)$, where $\HH_2(m)$ is the maximum possible
number of the congruent triangles in a point set of size $m$ in
$R^3$. Total time spent for pair $(q_1,q_2)$ then is
$O(n\HH_2)$. Since we use $O(\alpha n)$ pairs, the overall running time is
$O(\alpha n^2\HH_2)$. Agarwal and Sharir~\cite{AgSh} show that 
$\HH_2(m)\leq m^{5\over 3}g(m)$, where $g(m)$ is a very slowly growing 
function of $m$ of inverse-Ackermann type. \nnote{may be we should say a bit more explicitly
  what $g(m)$ is}
\begin{small}

\begin{table}[h]
\begin{tabular}{|l|l|l|}
\hline
Algorithm & Original running time & Improved running time\\\hline\hline
Pose Clustering (e.g. \cite{Olson97}) & $O(m^3n^3S(m))$ & $O(m^3n^2S(m))$\\ \hline
Alignment (e.g. \cite{Akutsu2}) & $O(m^3 + m \lambda^{3,2}(m,n))S(m)$ & $O(m^3n^2S(m))$ \\ \hline
GHT (e.g. \cite{Akutsu2})& $O(m^3S(m) + \lambda^{3,2}(m,n)
S(\lambda^{3,2}(m,n)))$ & $O(m^3S(m) + n^2 \HH_2(m))$\\ \hline
Geometric hashing (e.g. \cite{LaWo})& $O(m^4S(m) + n^4\HH_3(m))$ & $O(m^4S(m) + n^2\HH_3(m))$\\ \hline
{\bf This paper} &&$O(m^3S(m) + n^2 \HH_2(m))$ \\\hline
\end{tabular}
\caption{\label{voting-time-comp} \emph{Time complexity comparison of deterministic
algorithms for exact-$\alpha$-LCP in $\mathbb{R}^3$.
 $S(x)$ is the query time for
the dictionary of size $x$, which can be taken to be $O(\log x)$ or smaller; 
$\HH_2(m)$ is the maximum number of triangles spanned by $m$ points in $\mathbb{R}^3$ 
that are congruent to a given triangle, we have $\HH_2(m)\leq m^{5\over 3}g(m)$, where $g(m)$ is a  
very slowly growing inverse-Ackermann type   
function of $m$ \cite{AgSh}, and can be treated as constant for all
practical purposes; $\HH_3(m)$ is the maximum number of tetrahedrons spanned by $n$ points in $\mathbb{R}^3$ 
that are congruent to a given tetrahedron, we have  $\HH_3(m)=O(m^{2+\delta})$ for any $\delta>0$ \cite{AgSh}; 
$\lambda^{3,2}(m,n)=\tilde{O}(\min\{m^{1.8}n^3,m^{1.95}n^{2.68}
+m^{1.8875}n^{2.8}\})$\cite{Akutsu2}.}} 
\end{table}
\end{small}

As is often the case for algorithms for LCP, analysis involves
determining quantities such as $H_2(m)$, which is a difficult problem.  In the above table we have 
tried to give references for the first four algorithms including the tightest analyses  rather than the original 
sources. Note that our algorithm is simpler than the others in the first column 
which involve checking for congruent simplices in a dictionary.

\nnote{The last remark seems to be false.}
\nnote{I think in the above we should put $\min(\lambda^{3,2}(m,n),\lambda^{3,2}(n,m))$ in place of $\lambda^{3,2}(m,n)$ }
\nnote{Should we remark that in our sampling we did not do what Akutsu does, that is he checks only congruent triangles
(for alignment).  We can also do that and improve the running time, though the analysis in that case is an open problem.}

\section{Expander-based sampling}

While for the exact-$\alpha$-LCP the simple pigeonhole sampling served us well,
for the \tolerant-$\alpha$-LCP we do not know any such simple  
scheme for choosing pairs. The reason is that now we not only need
to guarantee that each large set contain some sampled pairs, but also
that each large set contain a sampled pair with large length
(diameter-pair) as needed for the application of
Lemma~\ref{lem:diametric-lemma} in the \DA algorithm. 
Our approach is based on expander graphs (see, e.g., \cite{AlSp}).   
Informally, expander graphs have linear number of edges but the edges
are ``well-spread'' in the sense that there is an edge between any two
sufficiently large disjoint subsets of vertices.  Let $G$
be an expander graph with $\QQ$ as its vertex set.  We show that for each $S
\subseteq \QQ$, if $|S|$ is not too small, then there is an edge
$(u,v)$ in $G$ such that $(u,v) \in S^2$ and $||u v||$ approximates
the diameter of $S$.   

By choosing the pairs for the \DA algorithm from the
edge set of $G$ (the rest of the algorithm is same as
before), we obtain a bicriteria -- distance and size --
approximation algorithm as stated in Theorem~\ref{thm:expander-based}
below.  We first give a few definitions and recall a result
about expander graphs that we will need to prove the correctness of
our algorithm.

\begin{definition}
Let $S$ be a finite set of points of $\mathbb{R}^r$ for $r \geq 1$,
and let $0 \leq k \leq n$. Define  
$\diameter{S}{k} = \min_{T:|T|=k} \ms{diameter}(S\setminus T).$
\end{definition}

That is, $\diameter{S}{k}$ is the {\em minimum} of the diameter of the
sets obtained by deleting $k$ points from $S$. Clearly, $\diameter{S}{0}=\ms{diameter}(S)$.

Let $U$ and $V$ be two disjoint subsets of vertices of a graph $G$. Denote by
$e(U,V)$ the set of edges in $G$ with one end in $U$ and the other in
$V$. 
We will make use of the following well-known theorem about the
eigenvalues of graphs (see, e.g.~\cite{KrSu}, for the proof and related background).
\begin{theorem} Let $G$ be a $d$-regular graph on $n$ vertices. 
Let $d = \lambda_1 \geq \lambda_2 \geq \ldots \geq \lambda_n$ 
be the eigenvalues of the adjacency matrix of $G$. Denote 
$\lambda = \max_{2 \leq i \leq n} |\lambda_i|.$
Then for every two disjoint subsets $U, W \subset V$,
\begin{equation} \label{eqn:expander}
\left||e(U,W)|-\frac{d|U||W|}{n}\right| \leq \lambda \sqrt{|U||W|}.
\end{equation}
\end{theorem}

\begin{corollary} \label{cor:expander:edge}
Let $U, W \subset V$ be two disjoint sets with $|U|=|W| >
\frac{\lambda n}{d}$. Then $G$ has an edge in $U \times W$. 
\end{corollary}
\begin{proof} 
It follows from (\ref{eqn:expander}) that if $\frac{d|U||W|}{n} >
\lambda \sqrt{|U||W|}$ then $|e(U,W)|>0$, and since  
$|e(U,W)|$ is integral, $|e(U,W)|\geq 1$. But the above condition is
clearly true if we take $U$ and $W$ as in the statement  
of Corollary~\ref{cor:expander:edge}.
\end{proof}

There are efficient constructions of graph families known
with $\lambda < 2 \sqrt{d}$ (see, e.g., \cite{AlSp}).  Let us call
such graphs \emph{good expander graphs}.  
We can now state our main result for this section.
\begin{theorem}\label{thm:expander-based}
For an $\alpha$-LCP instance $(\PP, \QQ)$ with $\LCP(\PP,\QQ)>\frac{n}{\alpha}$,
the \DA algorithm with expander-based sampling using a good
expander graph of degree $d>2500\alpha^2$ finds a rigid
motion $\mu$ in time $O(m^3n^2\log{m})$ such that there is a subset
$I$ satisfying the following criteria:
\begin{description}
\item{(1)} size-approximation criterion: 
    $|I| \ge \LCP(\PP,\QQ) - \frac{50}{\sqrt{d}}n$;
\item{(2)} distance-approximation criterion: each point of $\mu(I)$ is
within distance $6\epsilon$ from a point in $P$.
\end{description}
\end{theorem}
Thus by choosing $d$ large enough we can get as good size-approximation
as desired. 
The constants in the above theorem have been chosen for simplicity of
the proof and can be improved slightly.

For the proof we first need a lemma showing that choosing the query
pairs from a graph with small $\lambda(G)$ (the second largest
eigenvalue of $G$) gives a long (in a
well-defined sense) edge in every not too small subset of vertices.
\begin{lemma}\label{thm:expander:edge}
Let $G$ be a $d$-regular graph with vertex set $Q \subset \mathbb{R}^3$, and
$|Q|=n$. Let $S \subseteq Q$ be such that $|S|>\frac{25 \lambda(G)n}{d}$. 
Then there is an edge $\{s_1,s_2\} \in E(G)\cap S^2$ 
such that $||s_1s_2|| \geq {\diameter{S}{\frac{25 \lambda(G)}{d}n}\over 2}$. 
\end{lemma}
\begin{proof}
For a positive constant $c$ to be chosen later, remove $cn$ pairs from $S$ as follows. First remove a diameter 
pair, then from the remaining points remove a diameter pair, and so
on. Let $T$ be the set of points in the removed 
pairs and $T^p$ the set of removed pairs. 
The remaining set $S \setminus \TT$ has diameter $\geq
\diameter{S}{2cn}$ by the definition of $\diameter{S}{2cn}$, and hence each of the removed pairs
has length $\geq \diameter{S}{2cn}$. For $\BB, \CC \subset S$ let
$\dist{\BB}{\CC}  = \min_{b \in \BB, c \in \CC} ||bc||$.

\begin{claim}
The set $\TT$ defined above can be partitioned into three sets $\BB$, $\CC$,
$\EE$, such that $|\BB|, |\CC| \geq \frac{cn}{6}$, and $\dist{\BB}{\CC} \geq {\diameter{S}{2cn} \over 2}$. 
\end{claim}
\begin{proof}
Fix a Cartesian 
coordinate system and consider the projections of the pairs in $T^p$ on the
$x$-, $y$- ,and $z$-axes. It is easy to see that
for at least one of these axes, at least $cn\over 3$ pairs have
projections of length $\geq {\diameter{S}{2cn} \over \sqrt{3}}$.  
Suppose without loss of generality that this is the case for the
$x$-axis, and denote the set of projections of pairs  
on the $x$-axis with length $\geq {\diameter{S}{2cn}\over \sqrt{3}}$
by $T_x^p$, and the set of points in the pairs in $T_x^p$ by
$T_x$.  We have $|T_x|\geq 2cn/3$.  Now consider a {\em sliding window} $W$  
on the $x$-axis of length ${\diameter{S}{2cn} \over 2}$, initially at $-\infty$, and
slide it to $+\infty$. At any position of $W$, each pair in  
$T_x^p$ has at most $1$ point in $W$, as the length of any pair is
more than the length of $W$. Thus at any position, 
$W$ contains $\leq {|T_x^p|} = |T_x|/2$ points.  It is now easy to see by
a standard continuity argument that there is a 
position of $W$, call it $\bar{W}$, where there are 
$\geq {|T_x|\over 4} \geq {cn\over 6}$ points of $T_x$ both to the left and to  
the right of $\bar{W}$. 

Now, $\BB$ is defined to be the set of points in $\TT$ whose
projection is in $T_x$ and is to the left of $\bar{W}$;  
similarly $\CC$ is the set of points in $T$ whose projection is in $T_x$ and is to the right of $\bar{W}$. 
Clearly any two points, one from $\BB$ and the other from $\CC$, are ${\diameter{S}{2cn} \over 2}$-apart.
\end{proof}
Coming back to the proof of Lemma~\ref{thm:expander:edge},
the property that we need from the query-graph is that for any two disjoint sets
$\BB, \CC \subset S$ of size $\delta |S|$, where $\delta$ is a
small positive constant, the query-graph should have an edge in $\BB \times \CC$. 

By Corollary~\ref{cor:expander:edge} if $|\BB| \geq {cn\over 6} > \frac{\lambda n}{d}$, and 
$|\CC| \geq {cn\over 6} > \frac{\lambda n}{d}$, that is, if $c >
{6\lambda \over d}$, then $G$ has an edge in $\BB \times \CC$.  
Taking $c = \frac{12.5 \lambda}{d}$ completes the proof of
Lemma~\ref{thm:expander:edge}. 
\end{proof}

\noindent{\it Proof of Theorem~\ref{thm:expander-based}}.
If we take $G$ to be a good expander graph then
Lemma~\ref{thm:expander:edge} gives that $G$ has an edge of length
$\geq {\diameter{S}{\frac{50}{\sqrt{d}}n} \over 2}$. Let $S$ also be a
solution to \tolerant-LCP for input $(P,Q)$ with error parameter
$\epsilon>0$. We have that one of the sampled pairs has
length at least ${\diameter{S}{\frac{50}{\sqrt{d}}n} \over 2}$. Thus
applying an appropriate variant (replacing the diameter pair by the
sampled pair with large length as guaranteed by
Lemma~\ref{thm:expander:edge}) of Lemma~\ref{lem:diametric-lemma},  
we get a rigid motion $\mu$ such that there is a subset
$I$ satisfying the following:
\begin{description}
\item{(1)} $|I| \ge |S| - \frac{50}{\sqrt{d}}n$ for any $d>2500\alpha^2$;
\item{(2)} Each point of $I$ is within $6\epsilon  (=\epsilon +
\epsilon + 4\epsilon)$ of a point in $M$. \nnote{why is this $6\epsilon$?}
\eop
\end{description}

\section{Discussion}

We have presented a new practical algorithm for point pattern matching. 
Our \DA algorithm is the fastest known distance-approximation
algorithm for \tolerant-LCP, and is simple compared to other known
distance-approximation algorithms and heuristics which involve
6-dimensional clustering.  
Our analysis of \DA is not tight, and perhaps better bounds can
be obtained if the interpoint distance is greater than $\epsilon$ by a
sufficiently large constant factor.  

Our technique of pigeonhole sampling yields speed-ups for
all four popular algorithms and also the fastest
known deterministic algorithm for the exact-LCP.  Again, our
algorithms are simpler than the previous best algorithms.  
Akutsu et al.~\cite{Akutsu2} give a tighter analysis for GHT in terms
of the function $\lambda^{3,2}(m,n)$.  Our analysis of \DA (and GHT) with
pigeonhole sampling was based on $H_2(m)$.  Presumably, a better
analysis similar to the idea  in \cite{Akutsu2} is possible. 
 
Point pattern matching is of fundamental importance for computer vision
and structural bioinformatics. Indeed, this investigation stemmed 
from research in structural bioinformatics. Current software, which
uses either geometric hashing or generalized Hough transform, can
immediately benefit from this work. We have implemented a randomized
version of 
\DA for molecular common substructure detection and the results were reported
in ~\cite{ChGo3}.

{\bf Acknowledgment.}
We thank S. Muthukrishnan and Ali Shokoufandeh for the helpful
comments and advice. 

\newpage
\begin{appendix}
\centerline{\Large Appendix}

\section{Voting Algorithms for Exact-LCP}

In this appendix, we review and compare four popular algorithms for
  exact-LCP:  {\em pose clustering}, {\em alignment}, {\em generalized Hough
  transform}(GHT), and {\em geometric hashing}. 
These algorithms are all based on a {\em voting}
  idea and are sometimes confused in the literature. 
Please see Algorithms \ref{PoseC}, \ref{Alignment},
  \ref{GHT}, \ref{GH}) for a full description of the algorithms
  in 
  their generic form independent of the search data structure used.
In particular, geometric hashing algorithms need not  
use a hash-table as a search data structure.
We describe all the algorithms in terms of a dictionary of
objects (which are either transformations or a set of points and can be
ordered lexicographically). 
Denote the query time for
this dictionary by $S(x)+O(k)$ where $x$ is the size of the
  dictionary, and $k$ is the size of the output depending on the query. For
  example, if the dictionary is implemented by a search tree 
we have $S(x)=O(\log x)$.

Pose clustering and alignment are the basic methods.
GHT and geometric hashing can be regarded as their respective
efficient implementations. Efficiency is achieved by preprocessing of
the point sets  using their rigid motion invariant keys 
which speeds-up the searches.

In pose clustering, 
for each pair of triplets $(q_1, q_2, q_3) \in \QQ$ and
$(p_1, p_2, p_3) \in \PP$, we check if they are congruent. If they are then we
compute the rigid motion  $\mu$ such that $\mu(q_1, q_2, q_3)=(p_1, p_2,
p_3)$. We then cast one vote for $\mu$. The rigid motion which receives the
maximum number of votes corresponds to the maximum matching transformation
sought. The running time of pose clustering
is $O(m^3n^3S(m^3n^3))$ as the size of the dictionary of transformations
can be as large as $O(m^3n^3)$.

In alignment, for each pair of triplets $(q_1, q_2, q_3) \in \QQ$ and 
$(p_1, p_2, p_3) \in \PP$ we check if they are congruent. If they are then
we compute the rigid motion  $\mu$ such that $\mu(q_1, q_2, q_3)=(p_1, p_2,
p_3)$. 
Then we count the number of points in $\mu(\QQ)$ that coincide with points in
$\PP$. This number gives the number of votes the rigid motion $\mu$
gets. The rigid motion which receives the
maximum number of votes corresponds to the maximum matching transformation
sought. The running time is $O(m^3n^4S(m))$.

The difference between pose clustering and alignment is the voting space:
in pose clustering voting is done for transformations while in
alignment it is for bases (triplets of points). 
In both pose clustering and alignment algorithms, 
each possible triplet in $\QQ$ is compared with each possible triplet in $\PP$.
However, by representing each triplet with its rigid motion invariant key, only 
triplets with the same key (rigid motion invariant) are needed to be compared.
This provides an
efficient implementation. For example, the 
GHT algorithm is
an efficient implementation of pose clustering. Here we preprocess $\PP$ by
storing the triplets of points with the rigid motion invariant
keys in a dictionary. 
Now for each triplet $(q_1, q_2, q_3)$ in $\QQ$ we find congruent
triplets in $\PP$ by searching for the rigid motion invariant key for
$(q_1, q_2, q_3)$. The rest of the algorithm is the same as pose
clustering. Similarly the geometric hashing algorithm is an
efficient implementation of the alignment method.

GHT is faster than geometric hashing, however
geometric hashing has the advantage that algorithm can stop as soon as
it has found a good match. Depending on the application this gives
geometric hashing advantage over GHT.


As observed by Olson~\cite{Olson97} and Akutsu et al.~\cite{Akutsu2},
pose clustering and GHT can be further improved. This is because 
a $k$-matching transformation can be identified by matching $(k-2)$ bases
which match a common pair. We call this
version of the generalized Hough transform the {\em pair-based
  version}; it is described below in Algorithm \ref{reduction-HT}. 
Although the worst case time complexity of
the pair-based version and the original version are the same, 
this will serve as a basis for our new
scheme, called {\em \DA}. The pair-based version
also allows efficient random sampling of pairs
\cite{Olson97,Akutsu2}. 

\begin{singlespace}

\begin{algorithm}[H]
\caption{Pose Clustering}
\label{PoseC}
\begin{alginc}

\Procedure{Pose Clustering}{$\PP,\QQ$}
\State Initialize an empty dictionary $\DD$ of rigid motions; 
\For {each triplet $(q_1,q_2,q_3)$ of $\QQ$}
\For {each triplet $(p_1,p_2,p_3)$ of $\PP$}
\If {$(q_1,q_2,q_3)$ is congruent to
$(p_1,p_2,p_3)$,} 
\State Compute the rigid motion $\mu$ which matches $(q_1,q_2,q_3)$  to
$(p_1,p_2,p_3)$ ;
\State Search $\mu$ in the dictionary $\DD$;
\State If found, increase the votes of $\mu$; otherwise insert $\mu$
with one vote. 
\EndIf
\EndFor
\EndFor
\State Return the maximum vote rigid motion in $\DD$;
\EndProcedure
\end{alginc}
\end{algorithm}

\begin{algorithm}[H]
\caption{Alignment}\label{Alignment}
\begin{alginc}
\Procedure{Alignment}{$\PP,\QQ$}
\For {each triplet $(q_1,q_2,q_3)$ of $\QQ$}
\For {each triplet $(p_1,p_2,p_3)$ of $\PP$}
\State If $(q_1,q_2,q_3)$ is congruent to
$(p_1,p_2,p_3)$, compute the rigid motion $\mu$;
\State Vote = 0;
\Comment Vote is a local counter for the transformation $\mu$.
\For {each remaining point $q \in \QQ$ and $p \in \PP$ }
\State If $\mu(q)=p$, then increase Vote by 1;
\EndFor
\State Keep the maximum vote and its associated transformation;
\EndFor
\EndFor
\State Return the maximum vote transformation.
\EndProcedure
\end{alginc}
\end{algorithm}

\begin{algorithm}
\caption{The original version of generalized Hough transform.}\label{GHT}
\begin{alginc}
\Procedure{Preprocessing}{}
\For {each triplet $(p_1,p_2,p_3)$ of $\PP$}
\State Compute and insert the {\em rigid motion invariant} key for 
$(p_1,p_2,p_3)$ into a dictionary $\DF$;
\EndFor 
\EndProcedure

\Procedure{Recognition}{}
\State Initialize an empty  dictionary $\DS$ of rigid motions;
\For {each triplet $(q_1,q_2,q_3)$ of $\QQ$}
\State Compute and search the {\em rigid motion invariant} key for 
$(q_1,q_2,q_3)$ in the dictionary $\DF$;
\For {each entry $(p_1,p_2,p_3)$ found,} 
\State Compute the rigid
motion $\mu$ which matches $(q_1,q_2,q_3)$ to $(p_1,p_2,p_3)$;
\State Search $\mu$ in the dictionary $\DS$;
\State If found, increase the votes of $\mu$; otherwise insert $\mu$
with one vote into $\DS$; 
\EndFor
\EndFor
\State Return the maximum vote rigid motion in $\DS$;
\EndProcedure
\end{alginc}
\end{algorithm}

\begin{algorithm}[H]
\caption{Geometric Hashing}\label{GH}
\begin{alginc}
\Procedure{Preprocessing}{}
\For {each triplet $(p_1,p_2,p_3)$ of $\PP$}
\For {each of the remaining point $p$ of $\PP$}
\State Compute and insert the {\em rigid motion invariant} key for 
$\{(p_1,p_2,p_3),p\}$ into a dictionary $\DF$;
\EndFor
\EndFor 
\EndProcedure

\Procedure{Recognition}{}
\For {each triplet $(q_1,q_2,q_3)$ of $\QQ$}
\State Build an empty dictionary $\DS$ (of triplets of $\PP$);
\For {each of the remaining point $q$ of $\QQ$}
\State Compute and search the {\em rigid motion invariant} key for 
$\{(q_1,q_2,q_3),q\}$ in the dictionary $\DF$;
\For {each entry $\{(p_1,p_2,p_3),p\}$ found}
\State If $(p_1,p_2,p_3)$ exists in $\DS$, then increase its vote by
one; otherwise insert $(p_1,p_2,p_3)$ into $\DS$ with vote one.
\EndFor
\EndFor
\State Keep the maximum vote and compute the corresponding
transformation  from its associated triplet;
\EndFor 
\EndProcedure
\end{alginc}
\end{algorithm}

\begin{algorithm}[H]
\caption{The pair-based version of generalized Hough transform.}\label{GHT-2}
\label{reduction-HT}
\begin{alginc}
\Procedure{Preprocessing}{}
\For {each pair $(p_1,p_2)$ of $\PP$}
\For {each remaining point $p$ of $\PP$}
\State Compute and insert the {\em rigid motion invariant} key for 
$\{(p_1,p_2),p\}$ into a dictionary $\DD$;
\EndFor 
\EndFor 
\EndProcedure

\Procedure{Recognition}{}
\For {each pair $(q_1,q_2)$ of $\QQ$}
\State Initialize an empty  dictionary $\DS$ of rigid motions;
\For {each remaining point $q$ of $\QQ$}
\State Compute and search the {\em rigid motion invariant} key for 
$\{(q_1,q_2), q\}$ in the dictionary $\DD$;
\For {each entry $\{(p_1,p_2),p\}$ found,}
\State Compute the rigid
motion $\mu$ which matches $\{(q_1,q_2),q\}$ to $\{(p_1,p_2),p\}$. 
\State Search $\mu$ in the dictionary $\DS$;
\State If found, increase the votes of $\mu$; otherwise insert $\mu$
with one vote into $\DS$; 
\EndFor
\EndFor
\State Keep the rigid motion for $(q_1,q_2)$ that receives the
maximum number of votes.
\EndFor
\State Return the rigid motion that receives the maximum number of
votes among all pairs.
\EndProcedure
\end{alginc}
\end{algorithm}

\end{singlespace}

\end{appendix}


\begin{thebibliography}{10}
\bibitem{AgSh}
P. K. Agarwal and M. Sharir,
\newblock The number of congruent simplices in a point set, 
\newblock Discrete Comput. Geom. 28 no. 2 (2002) 123--150. 

\bibitem{AjMe}
M. Ajtai, N. Megiddo,
\newblock A Deterministic Poly(log log N)-Time N-Processor Algorithm
for Linear Programming in Fixed Dimension,
\newblock in: Proc. 24th ACM Symp. on Theory of Computing (1992), 327--338.

\bibitem{AlSp}
N.~Alon and J.~Spencer,
\newblock The Probabilistic Method,
\newblock (Wiley-Interscience, 2000)

\bibitem{Akutsu1}
T. Akutsu,
\newblock Protein structure alignment using dynamic programming and
iterative improvement,
\newblock IEICE Transactions on Information and Systems 12 (1996) 1629--1636.

\bibitem{Akutsu2}
T. Akutsu, H. Tamaki, T. Tokuyama,
\newblock Distribution of Distances and Triangles in a Point Set and
Algorithms for Computing the Largest Common Point Sets,
\newblock Discrete \& Computational Geometry 20 no.3 (1998) 307--331. 


\bibitem{Akutsu3}
T. Akutsu, K. Kanaya, A. Ohyama, A. Fujiyama,
\newblock Point matching under non-uniform distortion,
\newblock Discrete Applied Mathematics, 127(1) (2003) 5--21.

\bibitem{AltGuibas}
H. Alt and L.J. Guibas,
\newblock Discrete Geometric Shapes: Matching, Interpolation, and
Approximation,
\newblock in: J.-R. Sack, J. Urrutia, eds., Handbook of
Computational Geometry,  
\newblock (Elsevier Science Publishers B.V. North-Holland, Amsterdam,
1999) 121--153.

\bibitem{Alt88}
H. Alt, K. Mehlhorn, H. Wagener, E. Welzl,
\newblock Congruence, Similarity, and Symmetries of Geometric
Objects,
\newblock Discrete \& Computational Geometry 3 (1988) 237--256.

\bibitem{AmChGa}
C. Amb{\"u}hl, S. Chakraborty, B. G{\"a}rtner.
\newblock Computing Largest Common Point Sets under Approximate
Congruence.
\newblock ESA 2000, Lecture Notes in Computer Science 1879 Springer
2000: 52--63.

\bibitem{BiCh}
S. Biswas, S. Chakraborty.
\newblock Fast Algorithms for Determining Protein Structure Similarity.
\newblock Workshop on Bioinformatics and Computational Biology, at the International Conference on High Performance Computing (HiPC), Hyderabad, India, December 2001.


\bibitem{ChBi}
S. Chakraborty, S. Biswas.
\newblock Approximation Algorithms for 3-D Commom Substructure
Identification in Drug and Protein Molecules. 
\newblock Workshop on Algorithms and Data Structures (WADS),  1999.
Lecture Notes in Computer Science 1663.

\bibitem{CaSc}
D.~Cardoze, L.~Schulman.
\newblock Pattern Matching for Spatial Point Sets.
\newblock Proc. 39th FOCS 156--165, 1998.


\bibitem{Kedem}
L.P. Chew, D. Dor, A. Efrat and K. Kedem.
\newblock Geometric Pattern Matching in d-Dimensional Space, 
\newblock  Discrete and Computational Geometry, 21(1999), pp. 257-274. 

\bibitem{ChGo}
 V. Choi, N. Goyal.
\newblock A Combinatorial Shape Matching Algorithm for Rigid Protein
 Docking. 
\newblock Combinatorial Pattern Matching (CPM) 2004 Lecture Notes in
 Computer Science 3109 Springer 2004: 285-296. 


\bibitem{ChGo2}
V.~Choi, N.~Goyal.
\newblock An Efficient Approximation Algorithm for Point Pattern
Matching Under Noise. 
\newblock The 7th International Symposium,
  Latin American Theoretical Informatics (LATIN 2006). Valdivia, Chile, March 19--24, 2006.
 Lecture Notes in Computer Science, Vol. 3887, 2006, pp 298--310.

\bibitem{ChGo3}
V.~Choi, N.~Goyal.
\newblock
An Algorithmic Approach to the Identification  of Rigid Domains
  in Proteins. 
\newblock Submitted to 
Algorithmica's special issue on
  algorithms for processing protein structures.





\bibitem{Rapid97}
P. Finn, L. Kavraki, J-C. Latombe, R. Motwani,
C. Shelton, 
S. Venkatasubramanian, A. Yao.
\newblock RAPID: Randomized Pharmacophore Identification in Drug Design
\newblock The 13th Symposium on Computational Geometry, 1997. 
Computational Geometry: Theory and Applications 10 (4), 1998. 


\bibitem{Gavrilov}
M. Gavrilov, P. Indyk, R. Motwani, S.
Venkatasubramanian.
\newblock Combinatorial and Experimental Methods for Approximate Point
Pattern Matching. 
\newblock Algorithmica 38(1): 59--90 (2003). 



\bibitem{GoMiOr}
M.~T.~Goodrich, J. S.~B.~Mitchell, M. W.~Orletsky.
\newblock Approximate Geometric Pattern Matching Under Rigid Motions. 
\newblock IEEE Trans. Pattern Anal. Mach. Intell. 21(4): 371-379 (1999).


\bibitem{GrHu1}
W.~E.~L.~Grimson, D.~P.~Huttenlocher.
\newblock On the sensitivity of geometric hashing.
\newblock In \emph{Proceedings of the 3rd International Conference on
Computer Vision}: 334--338 (1990)

\bibitem{GrHu2}
W.~E.~L.~Grimson, D.~P.~Huttenlocher.
\newblock On the sensitivity of Hough transform for object
recognition.
\newblock IEEE Trans. on Pattern Analysis and Machine Intell. 12(3): 1990.


\bibitem{HeBo}
Y. Hecker and R. Bolle. 
\newblock On geometric hashing and the generalized Hough transform.
\newblock IEEE Trans. on Systems, Man and Cybernetics, 24, pp. 1328-1338, 1994

\bibitem{HeSc}
P. J. Heffernan, S. Schirra.
\newblock Approximate Decision Algorithms for Point Set Congruence,
\newblock Comput. Geom. 4 (1994) 137--156.


\bibitem{HK}
J. Hopcroft, R. Karp.
\newblock An $n^{5/2}$ algorithm for maximum matchings in bipartite graphs. 
\newblock SIAM J. Comp. 2 (1973) 225--231.


\bibitem{Indyk}
P. Indyk, R. Motwani, S.
Venkatasubramanian.
\newblock Geometric Matching under Noise: Combinatorial Bounds and Algorithms. 
\newblock SODA, 1999.

\bibitem{IrRa96}
S. Irani, P. Raghavan.
\newblock Combinatorial and experimental results for randomized point
matching algorithms.
\newblock The 12th Symposium on Computational Geometry, 1996.  Comput. Geom. 12(1-2): 17--31 (1999).

\bibitem{KaSh97}
M. Katz, M. Sharir.
\newblock An expander-based approach to geometric optimization. 
\newblock SIAM J. Comput., Vol. 26, No. 5, 1384--1408, 1997. 

\bibitem{Kenyon}
C. Kenyon, Y. Rabani, and A. Sinclair.
\newblock Low Distortion Maps Between Point Sets.
\newblock Proceedings of the Thirty-Sixth Annual ACM Symposium on Theory of Computing
     (STOC), 2004. 

\bibitem{KrSu}
M.~Krivelevich, B.~Sudakov.
\newblock Pseudo-random graphs.
\newblock Preprint. Available at {\tt http://www.math.tau.ac.il/\verb+~+krivelev/papers.html}

\bibitem{LaWo}
Y.~Lamdan, H.~J.~Wolfson.
\newblock Geometric Hashing: A general and efficient model-based
recognition scheme.
\newblock In \emph{Second International Conference on Computer
Vision}, 238--249 (1988).

\bibitem{Olson97}
C. F. Olson.
\newblock Efficient Pose Clustering Using a Randomized Algorithm.
\newblock International Journal of Computer Vision 23(2), 131--147 (1997).

\bibitem{WoRi}
H. J. Wolfson and I. Rigoutsos. 
\newblock Geometric hashing: an overview. 
\newblock IEEE Computational Science and Engineering, Vol 4, 10--21 (1997).




\bibitem{Wolfson2}
R.~Nussinov, H.J.~Wolfson.
\newblock Efficient Detection of Three - Dimensional Motifs In Biological
Macromolecules by Computer Vision Techniques,
\newblock Proc. of the Nat'l Academy of Sciences, {\bf 88} (1991)
10495 -- 10499.
\end{thebibliography}
\end{document}